\pdfoutput=1
\documentclass{article} 
\usepackage{iclr2017_conference,times}
\usepackage{subcaption}
\usepackage{graphicx} 
\usepackage{url}

\title{A Structured Self-attentive Sentence Embedding}

\begin{document}

\title{A Structured Self-attentive \\
Sentence Embedding}

\author{Zhouhan Lin$^{\ddagger\diamond }$\thanks{This work has been done during the 1st author's internship with IBM Watson.}, Minwei Feng$^\diamond $, Cicero Nogueira dos Santos$^\diamond $, Mo Yu$^\diamond $, \\
  {\bf Bing Xiang$^\diamond $, Bowen Zhou$^\diamond $ \& Yoshua Bengio$^{\ddagger\dag}$} \\
  $^\diamond $IBM Watson \\
  $^\ddagger$Montreal Institute for Learning Algorithms (MILA), Universit\'e de Montr\'eal \\
  $^\dag$CIFAR Senior Fellow \\
  \texttt{lin.zhouhan@gmail.com} \\
  \texttt{\{mfeng, cicerons, yum, bingxia, zhou\}@us.ibm.com}
}

%

\iclrfinalcopy 
\maketitle

\begin{abstract}
This paper proposes a new model for extracting an interpretable sentence
embedding by introducing self-attention. Instead of using a vector, we use a
2-D matrix to represent the embedding, with each row of the matrix attending on
a different part of the sentence. We also propose a self-attention mechanism
and a special regularization term for the model. As a side effect, the
embedding comes with an easy way of visualizing what specific parts of the
sentence are encoded into the embedding. We evaluate our model on 3 different
tasks: author profiling, sentiment classification and textual entailment.
Results show that our model yields a significant performance gain compared to
other sentence embedding methods in all of the 3 tasks.
\end{abstract}

\section{Introduction} \label{intro}

Much progress has been made in learning semantically meaningful distributed
representations of individual words, also known as word embeddings
\citep{bengio2001neural, mikolov2013efficient}. On the other hand, much remains
to be done to obtain satisfying representations of phrases
and sentences. Those methods generally fall into two categories. The first
consists of universal sentence embeddings usually
trained by unsupervised learning \citep{hill-cho-korhonen:2016:N16-1}. This
includes SkipThought vectors \citep{kiros2015skip}, ParagraphVector
\citep{le2014distributed}, recursive auto-encoders
\citep{socher-EtAl:2011:EMNLP, socher2013recursive}, Sequential Denoising
Autoencoders (SDAE), FastSent \citep{hill-cho-korhonen:2016:N16-1}, etc.

The other category consists of models trained specifically for a certain
task. They are usually combined with downstream
applications and trained by supervised learning. One generally finds that
specifically trained sentence embeddings perform better than generic ones,
although generic ones can be used in a semi-supervised setting, exploiting
large unlabeled corpora. Several models have been proposed along
this line, by using recurrent networks \citep{hochreiter1997long,
chung2014empirical}, recursive networks \citep{socher2013recursive} and
convolutional networks \citep{kalchbrenner2014convolutional, dossantos2014,
kim2014convolutional} as an intermediate step in creating sentence
representations to solve a wide variety of tasks including classification and
ranking \citep{yin2015convolutional, palangi2016deep, tanEtAl:2016, DBLP:conf/asru/FengXGWZ15}. 
A common approach in previous methods consists in creating a simple vector
representation by using the final hidden state of the RNN or the max (or
average) pooling from either RNNs hidden states or convolved n-grams.
Additional works have also been done in exploiting linguistic structures such as
parse and dependence trees to improve sentence representations
\citep{ma2015dependency, mou2015:EMNLP, tai-socher-manning:2015}.

For some tasks people propose to use attention mechanism on top of the CNN or LSTM model
to introduce extra source of information to guide the extraction of sentence
embedding \citep{santos2016attentive}. However, for some other tasks like
sentiment classification, this is not directly applicable since there is no such extra
information: the model is only given one single sentence as input. In those
cases, the most common way is to add a max pooling or averaging step across all
time steps \citep{lee2016sequential}, or just pick up the hidden representation
at the last time step as the encoded embedding \citep{margaritbatch}.

A common approach in many of the aforementioned methods consists of creating a
simple vector representation by using the final hidden state of the RNN or the
max (or average) pooling from either RNNs hidden states or convolved n-grams.
We hypothesize that carrying the semantics along all time steps of a recurrent
model is relatively hard and not necessary. We propose a self-attention
mechanism for these sequential models to replace the max pooling or averaging
step. Different from previous approaches, the proposed self-attention mechanism
allows extracting different aspects of the sentence into multiple vector
representations. It is performed on top of an LSTM in our sentence embedding
model. This enables attention to be used in those cases when there are no extra
inputs. In addition, due to its direct access to hidden representations from
previous time steps, it relieves some long-term memorization burden from LSTM. As a side
effect coming together with our proposed self-attentive sentence embedding,
interpreting the extracted embedding becomes very easy and explicit. 

Section \ref{approach} details on our proposed self-attentive sentence
embedding model, as well as a regularization term we proposed for this model,
which is described in Section \ref{reg}. We also provide a visualization method
for this sentence embedding in section \ref{visualization}. We then evaluate
our model in author profiling, sentiment classification and textual entailment
tasks in Section \ref{exps}.

\begin{figure}[!t]
    \centering
    \begin{subfigure}[!ht]{0.5\textwidth}
    \includegraphics[width=1\textwidth]{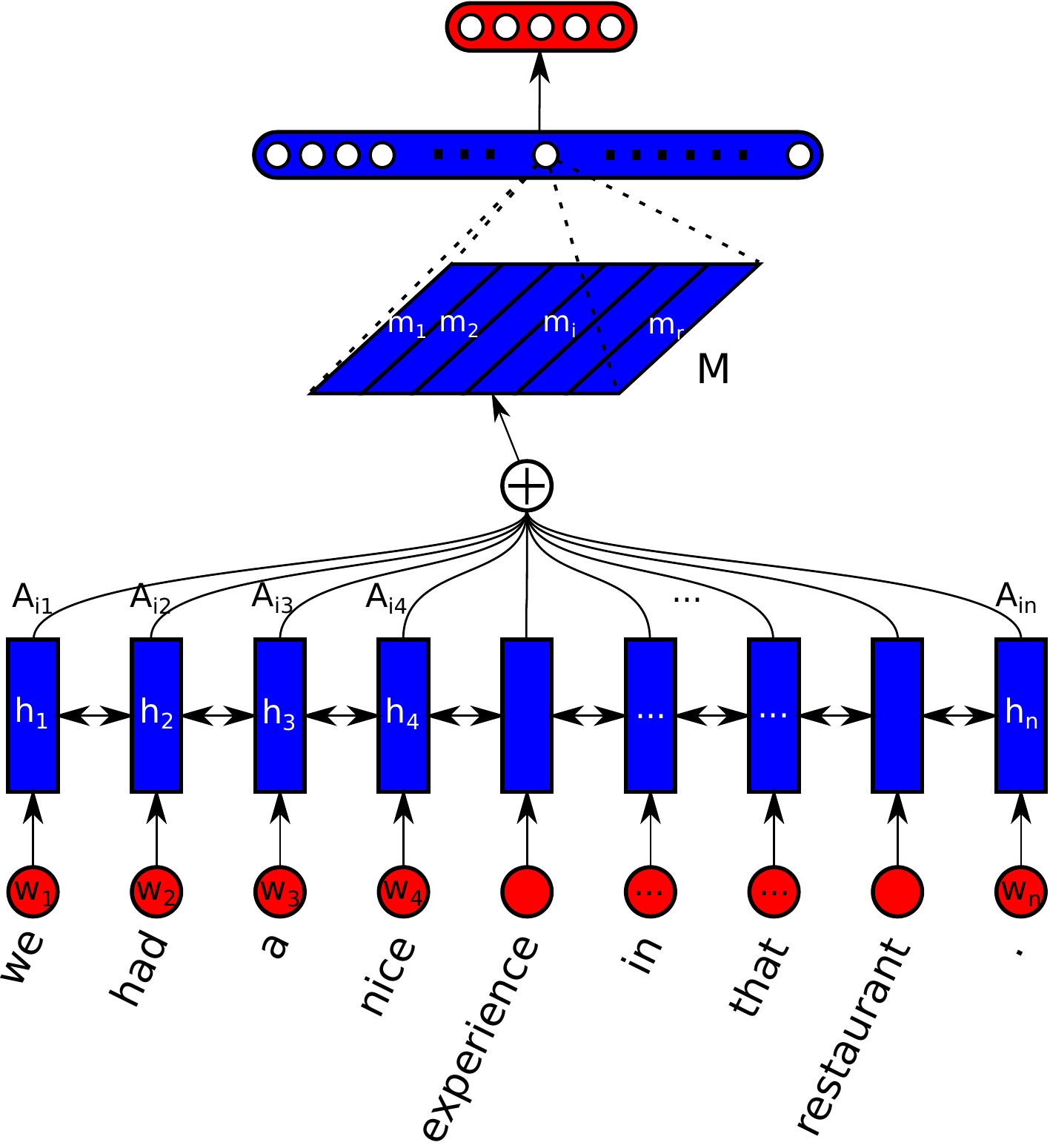}
    \caption{}
    \label{semlp}
    \end{subfigure}
    ~ 
    \begin{subfigure}[!ht]{0.25\textwidth}
    \includegraphics[width=1\textwidth]{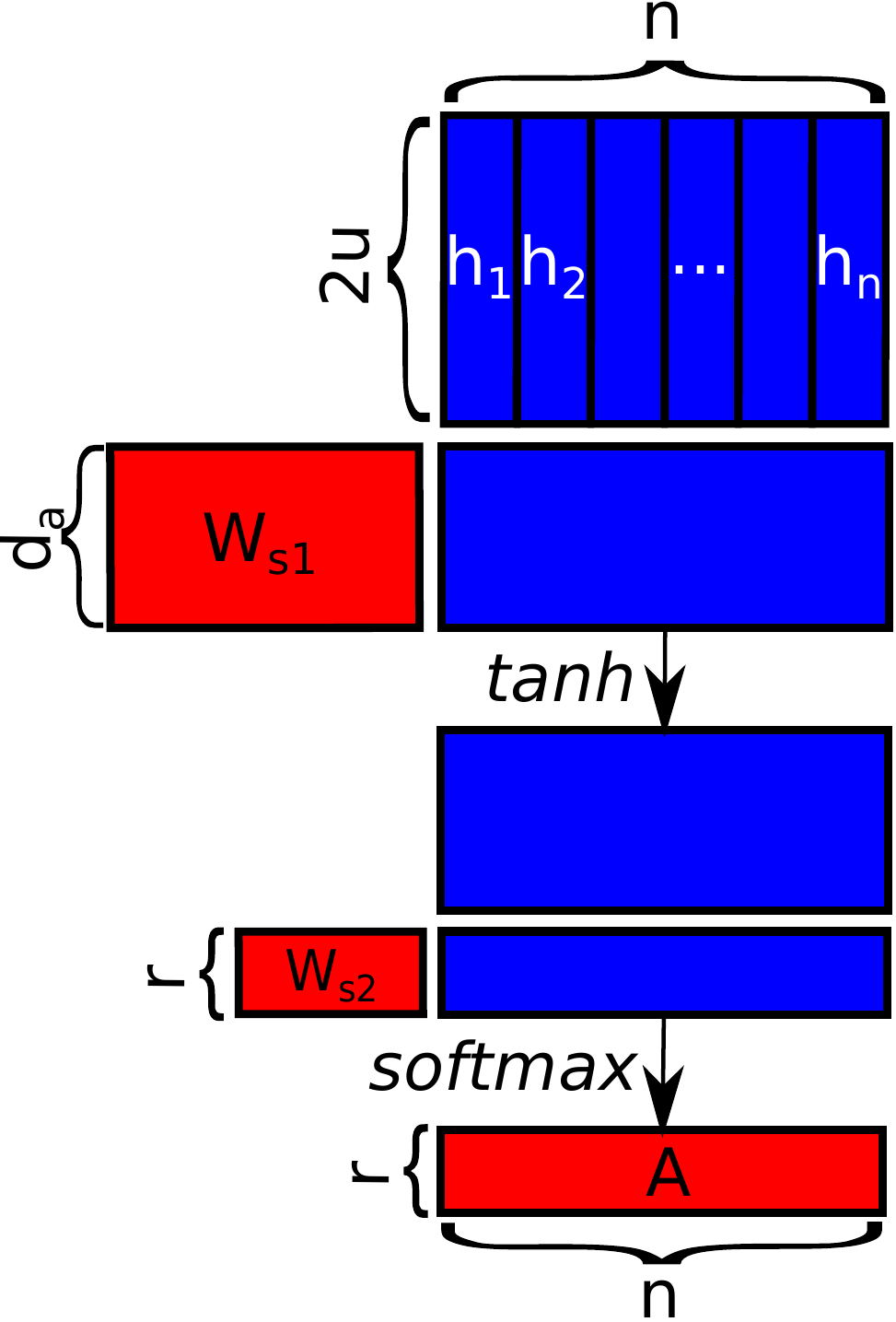}
    \caption{}
    \label{annotation}
    \end{subfigure}

    \caption{A sample model structure showing the sentence embedding model
combined with a fully connected and softmax layer for sentiment analysis (a).
The sentence embedding $M$ is computed as multiple weighted sums of hidden
states from a bidirectional LSTM ($\mathbf{h_1}, ..., \mathbf{h_n}$), where the
summation weights ($A_{i1}, ..., A_{in}$) are computed in a way illustrated in
(b). Blue colored shapes stand for hidden representations, and red colored
shapes stand for weights, annotations, or input/output.}
    \label{model_structure}
\end{figure}

\section{Approach} \label{approach}
\subsection{Model}
The proposed sentence embedding model consists of two parts.
The first part is a bidirectional LSTM, and
the second part is the self-attention mechanism, which provides a set of
summation weight vectors for the LSTM hidden states. These set of summation
weight vectors are dotted with the LSTM hidden states, and the resulting
weighted LSTM hidden states are considered as an embedding for the sentence.
It can be combined with, for example, a multilayer perceptron to be applied
on a downstream application. Figure \ref{model_structure} shows an example
when the proposed sentence embedding model is applied to sentiment analysis,
combined with a fully connected layer and a softmax layer. Besides using
a fully connected layer, we also proposes an approach that prunes weight
connections by utilizing the 2-D structure of matrix sentence embedding, which
is detailed in Appendix \ref{appena}. For this section, we will use Figure
\ref{model_structure} to describe our model.

Suppose we have a sentence, which has $n$ tokens, represented in a sequence of word embeddings.
\begin{equation}
    S=\left( \mathbf{w_1}, \mathbf{w_2}, \cdots \mathbf{w_n} \right)
\end{equation}
Here ${ w }_{ i }$ is a vector standing for a $d$ dimentional word embedding
for the $i$-th word in the sentence. $S$ is thus a sequence represented as a
2-D matrix, which concatenates all the word embeddings together. $S$ should
have the shape $n$-by-$d$.

Now each entry in the sequence $S$ are independent with each other. To gain
some dependency between adjacent words within a single sentence, we use a
bidirectional LSTM to process the sentence:
\begin{equation}
    \overrightarrow { { h }_{ t } } =\overrightarrow { LSTM } (w_{ t },\overrightarrow { { h }_{ t-1 } } )
\end{equation}
\begin{equation}
    \overleftarrow { { h }_{ t } } =\overleftarrow { LSTM } (w_{ t },\overleftarrow { { h }_{ t+1 } } )
\end{equation}
And we concatenate each $\overrightarrow { { h }_{ t } }$ with $\overleftarrow
{ { h }_{ t } }$ to obtain a hidden state $h_t$. Let the hidden unit number for
each unidirectional LSTM be $u$. For simplicity, we note all the $n$ $h_t$s as
$H$, who have the size $n$-by-$2u$.
\begin{equation}
    H=\left( \mathbf{h_1}, \mathbf{h_2},\cdots \mathbf{h_n} \right)
\end{equation}

Our aim is to encode a variable length sentence into a fixed size embedding. We
achieve that by choosing a linear combination of the $n$ LSTM hidden vectors in
$H$. Computing the linear combination requires the self-attention mechanism.
The attention mechanism takes the whole LSTM hidden states $H$ as input, and
outputs a vector of weights $\mathbf{a}$:
\begin{equation}
    \mathbf{a}=softmax\left( \mathbf{w_{s2}}tanh\left({ W }_{ s1 }H^T\right) \right)
\end{equation}
Here ${W}_{s1}$ is a weight matrix with a shape of $d_a$-by-$2u$. and
$\mathbf{w_{s2}}$ is a vector of parameters with size $d_a$, where $d_a$ is a
hyperparameter we can set arbitrarily. Since $H$ is sized $n$-by-$2u$, the
annotation vector $a$ will have a size $n$. the $softmax(\dot)$ ensures all the
computed weights sum up to 1. Then we sum up the LSTM hidden states $H$
according to the weight provided by $\mathbf{a}$ to get a vector representation
$\mathbf{m}$ of the input sentence. 

This vector representation usually focuses on a specific component of the
sentence, like a special set of related words or phrases. So it is expected to
reflect an aspect, or component of the semantics in a sentence. However, there
can be multiple components in a sentence that together forms the overall
semantics of the whole sentence, especially for long sentences. (For example,
two clauses linked together by an "and.") Thus, to represent the overall
semantics of the sentence, we need multiple $\mathbf{m}$'s that focus on
different parts of the sentence. Thus we need to perform multiple hops of
attention. Say we want $r$ different parts to be extracted from the sentence,
with regard to this, we extend the $\mathbf{{w}_{s2}}$ into a $r$-by-$d_a$
matrix, note it as $W_{s2}$, and the resulting annotation vector $\mathbf{a}$
becomes annotation matrix $A$. Formally,
\begin{equation} \label{attention}
    A=softmax\left( { W }_{ s2 }tanh\left({ W }_{ s1 }H^T\right) \right)
\end{equation}
Here the $softmax(\dot)$ is performed along the second dimension of its input.
We can deem Equation \ref{attention} as a 2-layer MLP without bias, whose
hidden unit numbers is $d_a$, and parameters are $\left\{W_{s2},
W_{s1}\right\}$.

The embedding vector $m$ then becomes an $r$-by-$2u$ embedding matrix $M$. We
compute the $r$ weighted sums by multiplying the annotation matrix $A$ and LSTM
hidden states $H$, the resulting matrix is the sentence embedding:
\begin{equation}
    M=AH
\end{equation}

\subsection{Penalization term} \label{reg}
The embedding matrix $M$ can suffer from redundancy problems if the attention
mechanism always provides similar summation weights for all the $r$ hops. Thus
we need a penalization term to encourage the diversity of summation weight
vectors across different hops of attention.

The best way to evaluate the diversity is definitely the Kullback – Leibler
divergence between any 2 of the summation weight vectors. However, we found
that not very stable in our case. We conjecture it is because we are
maximizing a set of KL divergence (instead of minimizing only one, which is the
usual case), we are optimizing the annotation matrix A to have a lot of
sufficiently small or even zero values at different softmax output units, and
these vast amount of zeros is making the training unstable. There is another
feature that KL doesn't provide but we want, which is, we want each individual
row to focus on a single aspect of semantics, so we want the probability mass in the
annotation softmax output to be more focused. but with KL penalty we can’t
encourage that.

We hereby introduce a new penalization term which overcomes the aforementioned
shortcomings.
Compared to the KL divergence penalization,
this term consumes only one third of the computation.
We use the dot product of $A$ and its transpose, subtracted by an identity
matrix, as a measure of redundancy.
\begin{equation} \label{penalization_term}
    P = {{ \left\| {\left( AA^T-I \right)} \right\|  }_{ F }}^2
\end{equation}
Here ${\left\| \bullet \right\|}_F$ stands for the Frobenius norm of a matrix.
Similar to adding an L2 regularization term, this penalization term $P$ will be
multiplied by a coefficient, and we minimize it together with the original
loss, which is dependent on the downstream application.

Let's consider two different summation vectors $\mathbf{a^i}$ and
$\mathbf{a^j}$ in $A$. Because of the softmax, all entries within any summation
vector in $A$ should sum up to 1. Thus they can be deemed as probability masses
in a discrete probability distribution. For any non-diagonal elements $a_{ij}
(i \neq j)$ in the $AA^T$ matrix, it corresponds to a summation over
elementwise product of two distributions:
\begin{equation}
    0 < a_{ij} = \sum _{ k=1 }^{ n }{ { a }_{ k }^{ i }{ a }_{ k }^{ j } } < 1
\end{equation}
where $a_k^i$ and $a_k^j$ are the $k$-th element in the $\mathbf{a^i}$ and
$\mathbf{a^j}$ vectors, respectively. In the most extreme case, where there is
no overlap between the two probability distributions $\mathbf{a^i}$ and
$\mathbf{a^j}$, the correspond $a_{ij}$ will be 0. Otherwise, it will have a
positive value. On the other extreme end, if the two distributions are
identical and all concentrates on one single word, it will have a maximum value
of 1. We subtract an identity matrix from $AA^T$ so that forces the elements on
the diagonal of $AA^T$ to approximate 1, which encourages each summation vector
$\mathbf{a^i}$ to focus on as few number of words as possible, forcing each
vector to be focused on a single aspect, and all other
elements to 0, which punishes redundancy between different summation vectors.

\subsection{Visualization} \label{visualization}
The interpretation of the sentence embedding is quite straight forward because
of the existence of annotation matrix $A$. For each row in the sentence
embedding matrix $M$, we have its corresponding annotation vector
$\mathbf{a^i}$. Each element in this vector corresponds to how much
contribution the LSTM hidden state of a token on that position contributes to.
We can thus draw a heat map for each row of the embedding matrix $M$
This way of visualization gives hints on what is encoded in each part of the
embedding, adding an extra layer of interpretation. (See Figure
\ref{age_nopnty_extend} and \ref{age_pnty_extend}).

The second way of visualization can be achieved by summing up over all the
annotation vectors, and then normalizing the resulting weight vector to sum up
to 1. Since it sums up all aspects of semantics of a sentence, it yields a
general view of what the embedding mostly focuses on. We can figure out which
words the embedding takes into account a lot, and which ones are skipped by the
embedding. See Figure \ref{age_nopnty} and \ref{age_pnty}.

\section{Related work}

Various supervised and unsupervised sentence embedding models have been
mentioned in Section \ref{intro}.
Different from those models, our proposed method uses a
new self-attention mechanism that allows it to extract different aspects of the
sentence into multiple vector-representations. The matrix structure together
with the penalization term gives our model a greater capacity to disentangle
the latent information from the input sentence. We also do not use linguistic
structures to guide our sentence representation model. Additionally, using our
method we can easily create visualizations that can help in the interpretation
of the learned representations.

Some recent work have also proposed supervised methods that use
intra/self-sentence attention. \cite{ling2015not} proposed an attention based model
for word embedding, which calculates an attention weight for each word at each
possible position in the context window. However this method cannot be extended
to sentence level embeddings since one cannot exhaustively enumerate all
possible sentences. \cite{LiuSLW16} proposes a sentence level attention which
has a similar motivation but done differently. They utilize the mean pooling
over LSTM states as the attention source, and use that to re-weight the pooled
vector representation of the sentence. 

Apart from the previous 2 variants, we want to note that \cite{li2016dataset}
proposed a
same self attention mechanism for question encoding in their factoid QA model,
which is concurrent to our work. The difference lies in that their encoding is
still presented as a vector, but our attention produces a matrix representation
instead, with a specially designed penalty term. We applied the model for
sentiment anaysis and entailment, and their model is for factoid QA.

The LSTMN model \citep{cheng2016long} also proposed a very successful
intra-sentence level attention mechanism, which is later used by
\cite{Parikh2016}. We see our attention and theirs as having different
granularities. LSTMN produces an attention vector for each of its hidden states
during the recurrent iteration, which is sort of an "online updating"
attention. It's  more fine-grained, targeting at discovering lexical
correlations between a certain word and its previous words. On the contrary, our
attention mechanism is only performed once, focuses directly on the semantics
that makes sense for discriminating the targets. It is less focused on
relations between words, but more on the semantics of the whole sentence that
each word contributes to. Computationally, our method also scales up with the
sentence length better, since it doesn't require the LSTM to compute an
annotation vector over all of its previous words each time when the LSTMN
computes its next step.



\section{Experimental results}  \label{exps}
We first evaluate our sentence embedding model by applying it to 3 different
datasets: the Age dataset, the Yelp dataset, and the Stanford Natural Language
Inference (SNLI) Corpus. These 3 datasets fall into 3 different tasks,
corresponding to author profiling, sentiment analysis, and textual entailment,
respectively. Then we also perform a set of exploratory experiments to validate
properties of various aspects for our sentence embedding model.

\subsection{Author profiling}  \label{age}
The Author Profiling
dataset\footnote{\texttt{http://pan.webis.de/clef16/pan16-web/author-profiling.html}}
consists of Twitter tweets in English, Spanish, and Dutch. For some of the
tweets, it also provides an age and gender of the user when writing the tweet.
The age range are split into 5 classes: 18-24, 25-34, 35-49, 50-64, 65+. We use
English tweets as input, and use those tweets to predict the age range of the
user. Since we are predicting the age of users, we refer to it as Age dataset
in the rest of our paper. We randomly selected 68485 tweets as training
set, 4000 for development set, and 4000 for test set. Performances are also
chosen to be classification accuracy.

\begin{table}[h]
\caption{Performance Comparision of Different Models on Yelp and Age Dataset}
\label{yelpage}
\begin{center}
\begin{tabular}{lll}
\multicolumn{1}{c}{\bf Models}  &\multicolumn{1}{c}{\bf Yelp}  &\multicolumn{1}{c}{\bf Age}
\\ \hline
BiLSTM + Max Pooling + MLP      & 61.99\%                      & 77.40\%          \\
CNN + Max Pooling + MLP         & 62.05\%                      & 78.15\%          \\
Our Model                       & {\bf 64.21\%}                & {\bf 80.45\%} \\
\end{tabular}
\end{center}
\end{table}

We compare our model with two baseline models: biLSTM and CNN. 
For the two baseline models. The biLSTM model uses a bidirectional LSTM with
300 dimensions in each direction, and use max pooling across all LSTM hidden
states to get the sentence embedding vector, then use a 2-layer ReLU output MLP
with 3000 hidden states to output the classification result. The CNN model uses
the same scheme, but substituting biLSTM with 1 layer of 1-D convolutional
network. During training we use 0.5 dropout on the MLP and 0.0001 L2
regularization. We use stochastic gradient descent as the optimizer, with a
learning rate of 0.06, batch size 16. For biLSTM, we also clip the norm of
gradients to be between -0.5 and 0.5. We searched hyperparameters in a wide
range and find the aforementioned set of hyperparameters yields the highest
accuracy.

For our model, we use the same settings as what we did in biLSTM. We also use a
2-layer ReLU output MLP, but with 2000 hidden units. In addition, our
self-attention MLP has a hidden layer with 350 units (the $d_a$ in Section
\ref{approach}), we choose the matrix embedding to have 30 rows (the $r$), and
a coefficient of 1 for the penalization term. 

We train all the three models until convergence and select the corresponding
test set performance according to the best development set performance. Our
results show that the model outperforms both of the biLSTM and CNN baselines by
a significant margin.

\subsection{Sentiment analysis}

\begin{figure}[t]
    \centering
    \begin{subfigure}[!hb]{0.8\textwidth}
    \includegraphics[width=1\textwidth]{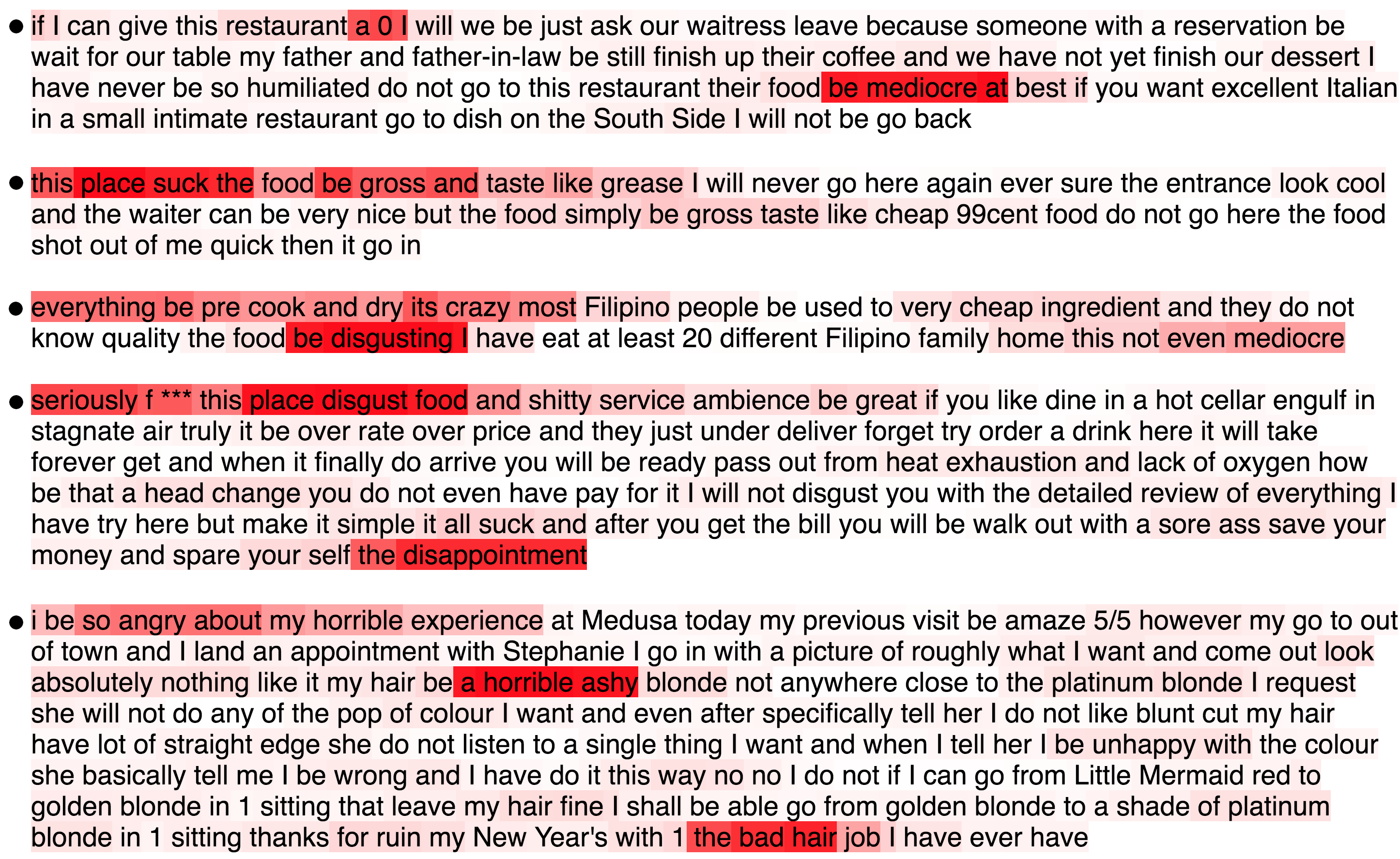}
    \caption{1 star reviews}
    \label{yelp_1star}
    \end{subfigure}
    ~ 
    \begin{subfigure}[!hb]{0.8\textwidth}
    \includegraphics[width=1\textwidth]{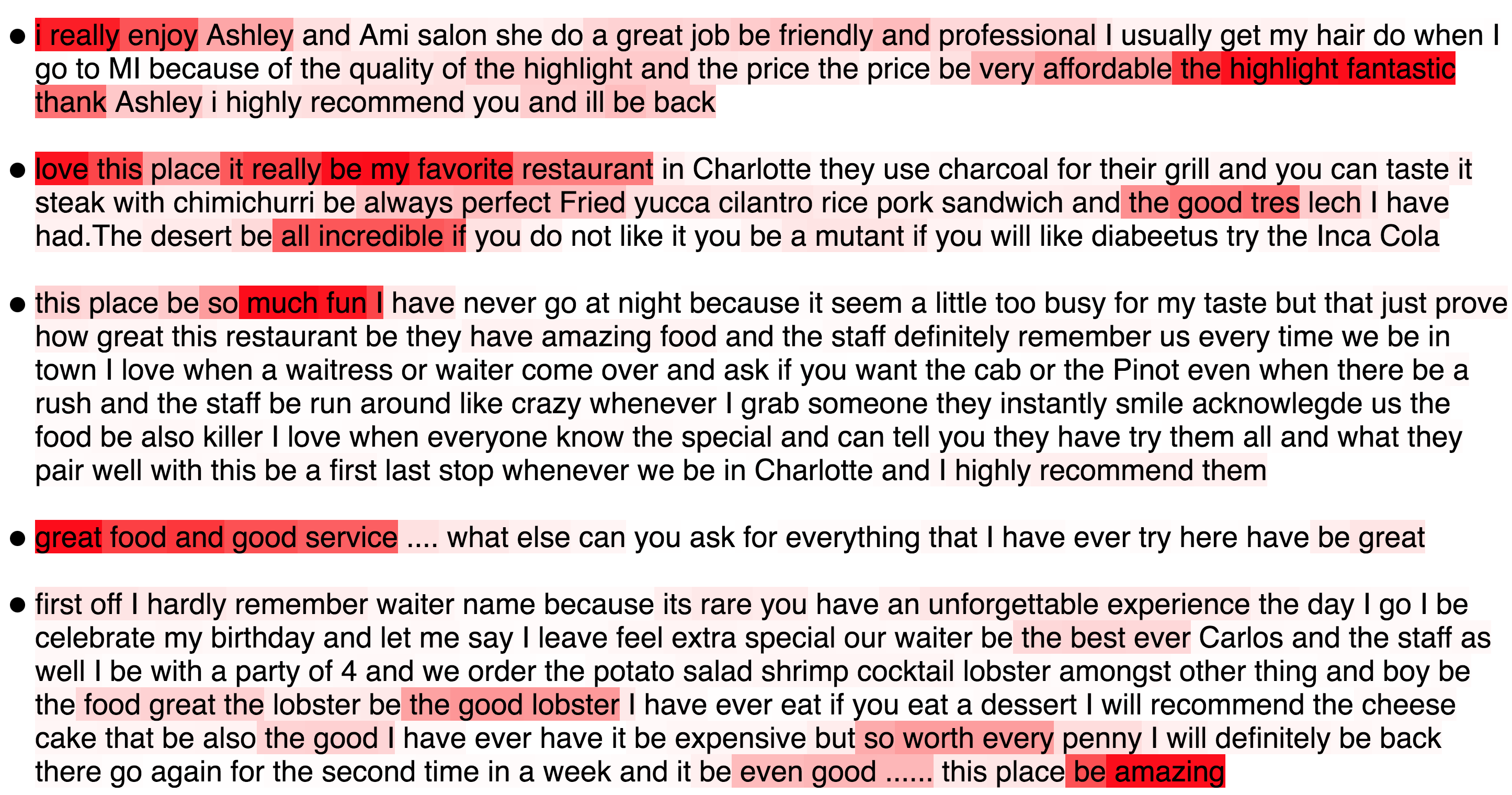}
    \caption{5 star reviews}
    \label{yelp_5star}
    \end{subfigure}
    \caption{Heatmap of Yelp reviews with the two extreme score.}
    \label{yelp_reviews}
\end{figure}

We choose the Yelp
dataset\footnote{\texttt{https://www.yelp.com/dataset\_challenge}} for
sentiment analysis task. It consists of 2.7M yelp reviews, we take the review
as input and predict the number of stars the user who wrote that review
assigned to the corresponding business store. We randomly select 500K
review-star pairs as training set, and 2000 for development set, 2000 for test
set. We tokenize the review texts by Stanford tokenizer. We use 100 dimensional
word2vec as initialization for word embeddings, and tune the embedding during
training across all of our experiments. The target number of stars is an
integer number in the range of $[1, 5]$, inclusive. We are treating the task as
a classification task, i.e., classify a review text into one of the 5 classes.
We use classification accuracy as a measurement.

For the two baseline models, we use the same setting as what we used for Author
Profiling dataset, except that we are using a batch size of 32 instead. For our
model, we are also using the same setting, except that we choose the hidden
unit numbers in the output MLP to be 3000 instead. We also observe a
significant performance gain comparining to the two baselines. (Table
\ref{yelpage})

As an interpretation of the learned sentence embedding, we use the second way
of visualization described in Section \ref{visualization} to plot heat maps for
some of the reviews in the dataset. We randomly select 5 examples of negative
(1 star) and positive (5 stars) reviews from the test set, when the model has a
high confidence ($>0.8$) in predicting the label. As shown in Figure
\ref{yelp_reviews}, we find that the model majorly learns to capture some key
factors in the review that indicate strongly on the sentiment behind the
sentence. For most of the short reviews, the model manages to capture all the
key factors that contribute to an extreme score, but for longer reviews, the
model is still not able to capture all related factors. For example, in the 3rd
review in Figure \ref{yelp_5star}), it seems that a lot of focus is spent on
one single factor, i.e., the "so much fun", and the model puts a little amount
of attention on other key points like "highly recommend", "amazing food", etc.

\begin{table}[b]
    \caption{Test Set Performance Compared to other Sentence Encoding Based
             Methods in SNLI Datset}
    \label{snli_table}
    \begin{center}
    \begin{tabular}{ll}
    \multicolumn{1}{c}{\bf Model}  &\multicolumn{1}{c}{\bf Test Accuracy}
    \\ \hline
    300D LSTM encoders \citep{bowman2016fast}                      & 80.6\%  \\
    600D (300+300) BiLSTM encoders \citep{liu2016learning}         & 83.3\%  \\
    300D Tree-based CNN encoders \citep{mou2015natural}            & 82.1\%  \\
    300D SPINN-PI encoders \citep{bowman2016fast}                  & 83.2\%  \\
    300D NTI-SLSTM-LSTM encoders \citep{munkhdalai2016neural}      & 83.4\%  \\
    1024D GRU encoders with SkipThoughts pre-training \citep{vendrov2015order}  & 81.4\%  \\
    300D NSE encoders \citep{munkhdalai2016neuralb}                       & {\bf 84.6\%}  \\
    \hline
    Our method                              & 84.4\%  \\
    \end{tabular}
    \end{center}
\end{table}

\subsection{Textual entailment}  \label{snli}
We use the biggest dataset in textual entailment, the SNLI corpus
\citep{bowman2015large} for our evaluation on this task. SNLI is a collection
of 570k human-written English sentence pairs manually labeled for balanced
classification with the labels entailment, contradiction, and neutral. The
model will be given a pair of sentences, called hypothesis and premise
respectively, and asked to tell if the semantics in the two sentences are
contradicting with each other or not. It is also a classification task, so we
measure the performance by accuracy.

We process the hypothesis and premise independently, and then extract the
relation between the two sentence embeddings by using multiplicative interactions
proposed in \cite{memisevic2013learning} (see Appendix \ref{appenb} for details),
and use a 2-layer ReLU output MLP with 4000
hidden units to map the hidden representation into classification results.
Parameters of biLSTM and attention MLP are shared across hypothesis and
premise. The biLSTM is 300 dimension in each direction, the attention MLP has
150 hidden units instead, and both sentence embeddings for hypothesis and
premise have 30 rows (the $r$). The penalization term coefficient is set to
0.3. We use 300 dimensional GloVe \citep{pennington2014glove} word embedding to
initialize word embeddings. We use AdaGrad as the optimizer, with a learning
rate of 0.01. We don't use any extra regularization methods, like dropout or L2
normalization. Training converges after 4 epochs, which is relatively fast.

This task is a bit different from previous two tasks, in that it has 2
sentences as input. There are a bunch of ways to add inter-sentence level
attention, and those attentions bring a lot of benefits. To make the comparison
focused and fair, we only compare methods that fall into the sentence
encoding-based models. i.e., there is no information exchanged between the
hypothesis and premise before they are encoded into some distributed encoding.

We find that compared to other published approaches, our method shows a
significant gain ($\ge 1\%$) to them, except for the 300D NSE encoders, which
is the state-of-the-art in this category. However, the $0.2\%$ different is
relatively small compared to the differences between other methods.

\subsection{Exploratory experiments}
In this subsection we are going to do a set of exploratory experiments to study
the relative effect of each component in our model. 

\subsubsection{Effect of penalization term}
Since the purpose of introducing the penalization term $P$ is majorly to discourage
the redundancy in the embedding, we first directly visualize the heat maps of
each row when the model is presented with a sentence. We compare two identical
models with the same size as detailed in Section \ref{age} trained separately
on Age dataset, one with this penalization term (where the penalization
coefficient is set to 1.0) and the other with no penalty. We randomly select
one tweet from the test set and compare the two models by plotting a heat map
for each hop of attention on that single tweet. Since there are 30 hops of
attention for each model, which makes plotting all of them quite redundant, we
only plot 6 of them. These 6 hops already reflect the situation in all of the
30 hops.

\begin{figure}[!b]
    \centering
    \begin{subfigure}[!hb]{0.49\textwidth}
    \includegraphics[width=1\textwidth]{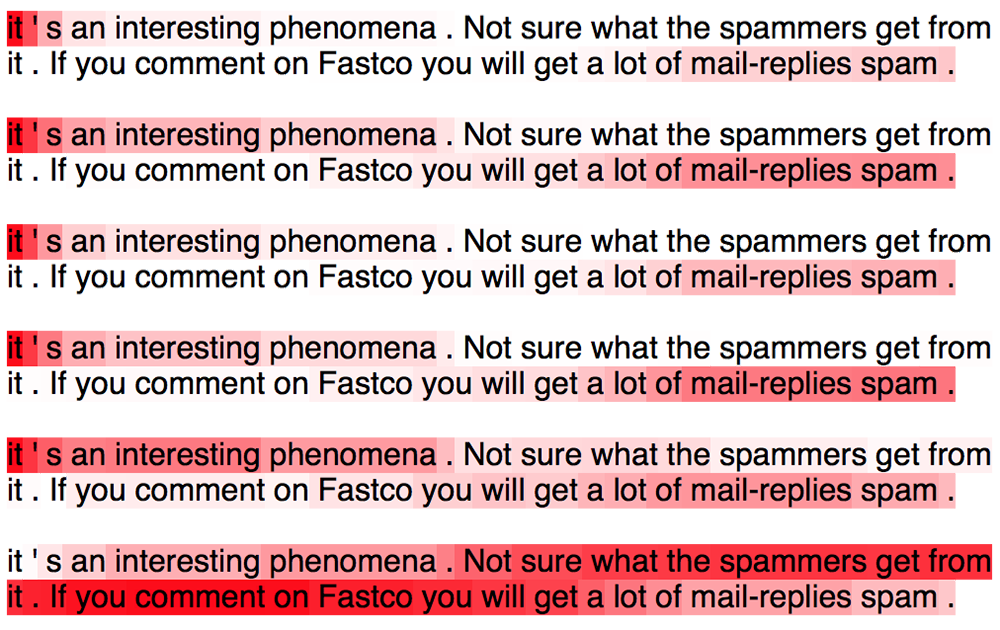}
    \caption{}
    \label{age_nopnty_extend}
    \end{subfigure}
    ~ 
    \begin{subfigure}[!hb]{0.49\textwidth}
    \includegraphics[width=1\textwidth]{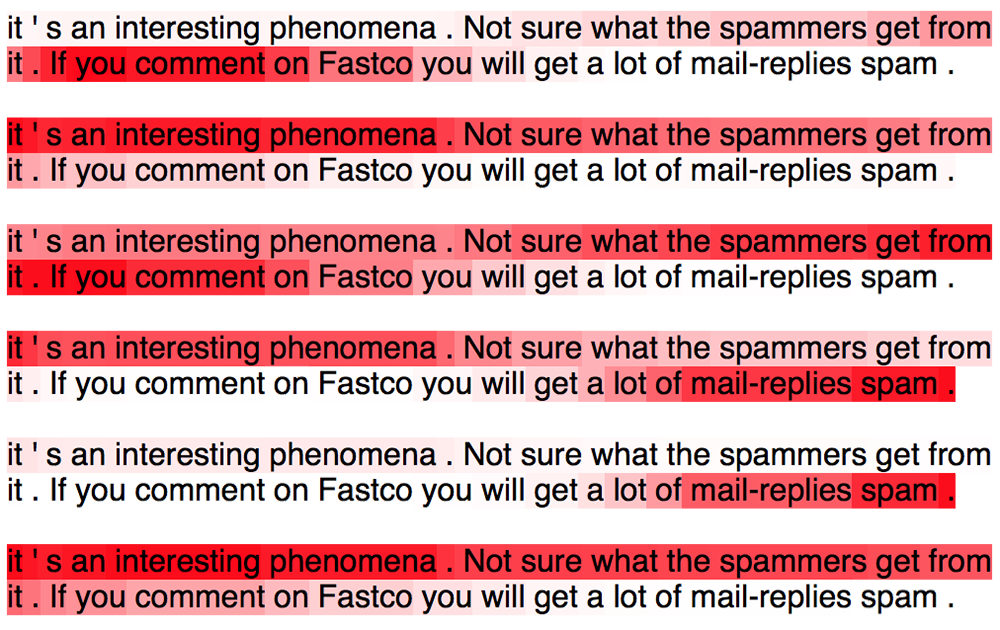}
    \caption{}
    \label{age_pnty_extend}
    \end{subfigure}

    \begin{subfigure}[!hb]{0.49\textwidth}
    \includegraphics[width=1\textwidth]{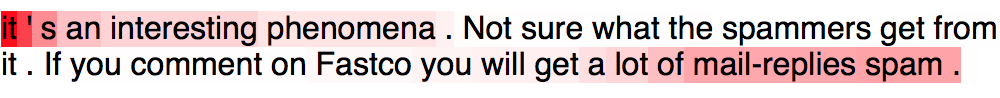}
    \caption{without penalization}
    \label{age_nopnty}
    \end{subfigure}
    ~ 
    \begin{subfigure}[!hb]{0.49\textwidth}
    \includegraphics[width=1\textwidth]{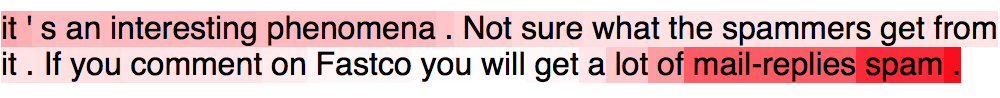}
    \caption{with 1.0 penalization}
    \label{age_pnty}
    \end{subfigure}
    
    \caption{Heat maps for 2 models trained on Age dataset. The left column is
    trained without the penalization term, and the right column is trained with
    1.0 penalization. (a) and (b) shows detailed attentions taken by 6 out of 30
    rows of the matrix embedding, while (c) and (d) shows the overall attention by
    summing up all 30 attention weight vectors. }
    \label{age_visualization_extended}
\end{figure}

\begin{figure}[!b]
    \centering
    \begin{subfigure}[!hb]{0.49\textwidth}
        \includegraphics[width=\linewidth]{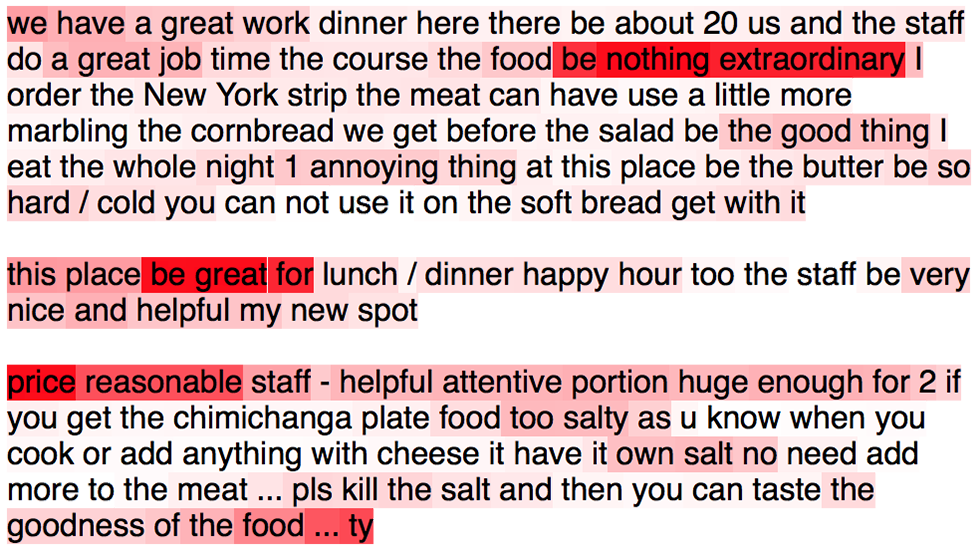}
        \caption{Yelp without penalization}
        \label{yelp_nopnty}
    \end{subfigure}
    ~ 
    \begin{subfigure}[!hb]{0.49\textwidth}
        \includegraphics[width=\linewidth]{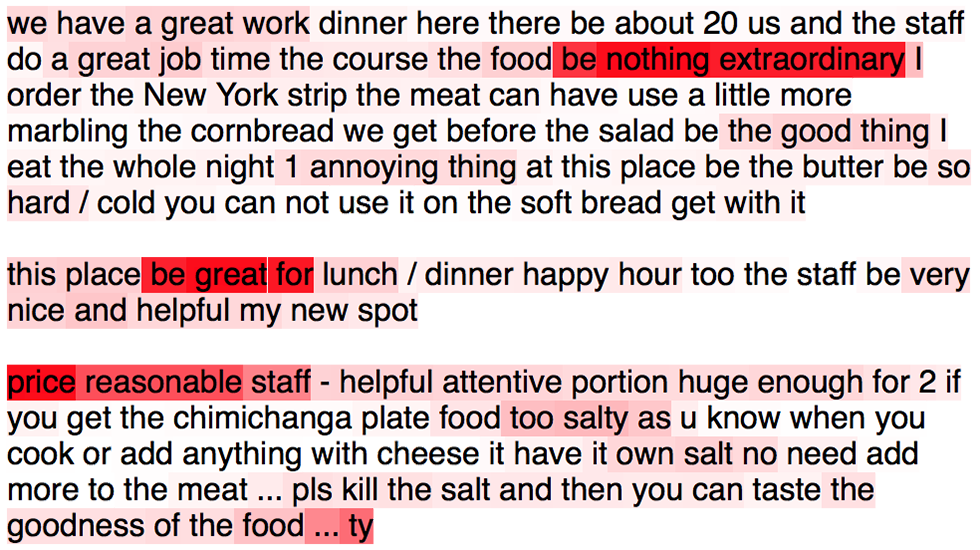}
        \caption{Yelp with penalization}
        \label{yelp_pnty}
    \end{subfigure}
    \caption{Attention of sentence embedding on 3 different Yelp reviews. The
    left one is trained without penalization, and the right one is trained with
    1.0 penalization.}
    \label{yelp_visualize}
\end{figure}

From the figure we can tell that the model trained without the penalization
term have lots of redundancies between different hops of attention (Figure
\ref{age_nopnty_extend}), resulting in putting lot of focus on the word "it"
(Figure \ref{age_nopnty}), which is not so relevant to the age of the author.
However in the right column, the model shows more variations between different
hops, and as a result, the overall embedding focuses on "mail-replies spam"
instead. (Figure \ref{age_pnty})

For the Yelp dataset, we also observe a similar phenomenon. To make the
experiments more explorative, we choose to plot heat maps of overall attention
heat maps for more samples, instead of plotting detailed heat maps for a single
sample again. Figure \ref{yelp_visualize} shows overall focus of the sentence
embedding on three different reviews. We observe that with the penalization
term, the model tends to be more focused on important parts of the review. We
think it is because that we are encouraging it to be focused, in the diagonals
of matrix $AA^T$ (Equation \ref{penalization_term}).

To validate if these differences result in performance difference, we evaluate
four models trained on Yelp and Age datasets, both with and without the
penalization term. Results are shown in Table \ref{yelpage_penalization}.
Consistent with what expected, models trained with the penalization term
outperforms their counterpart trained without.

\begin{table}[t]
\caption{Performance comparision regarding the penalization term}
\label{yelpage_penalization}
\begin{center}
\begin{tabular}{lll}
\multicolumn{1}{c}{\bf Penalization coefficient}  &\multicolumn{1}{c}{\bf Yelp}  &\multicolumn{1}{c}{\bf Age}
\\ \hline
1.0         & 64.21\%                      & 80.45\%          \\
0.0         & 61.74\%                      & 79.27\%          \\
\end{tabular}
\end{center}
\end{table}

In SNLI dataset, although we observe that introducing the penalization term
still contributes to encouraging the diversity of different rows in the matrix
sentence embedding, and forcing the network to be more focused on the
sentences, the quantitative effect of this penalization term is not so obvious
on SNLI dataset. Both models yield similar test set accuracies.

\subsubsection{Effect of multiple vectors}
\begin{figure}[!b]
    \centering
    \begin{subfigure}[]{0.49\textwidth}
        \includegraphics[width=\linewidth]{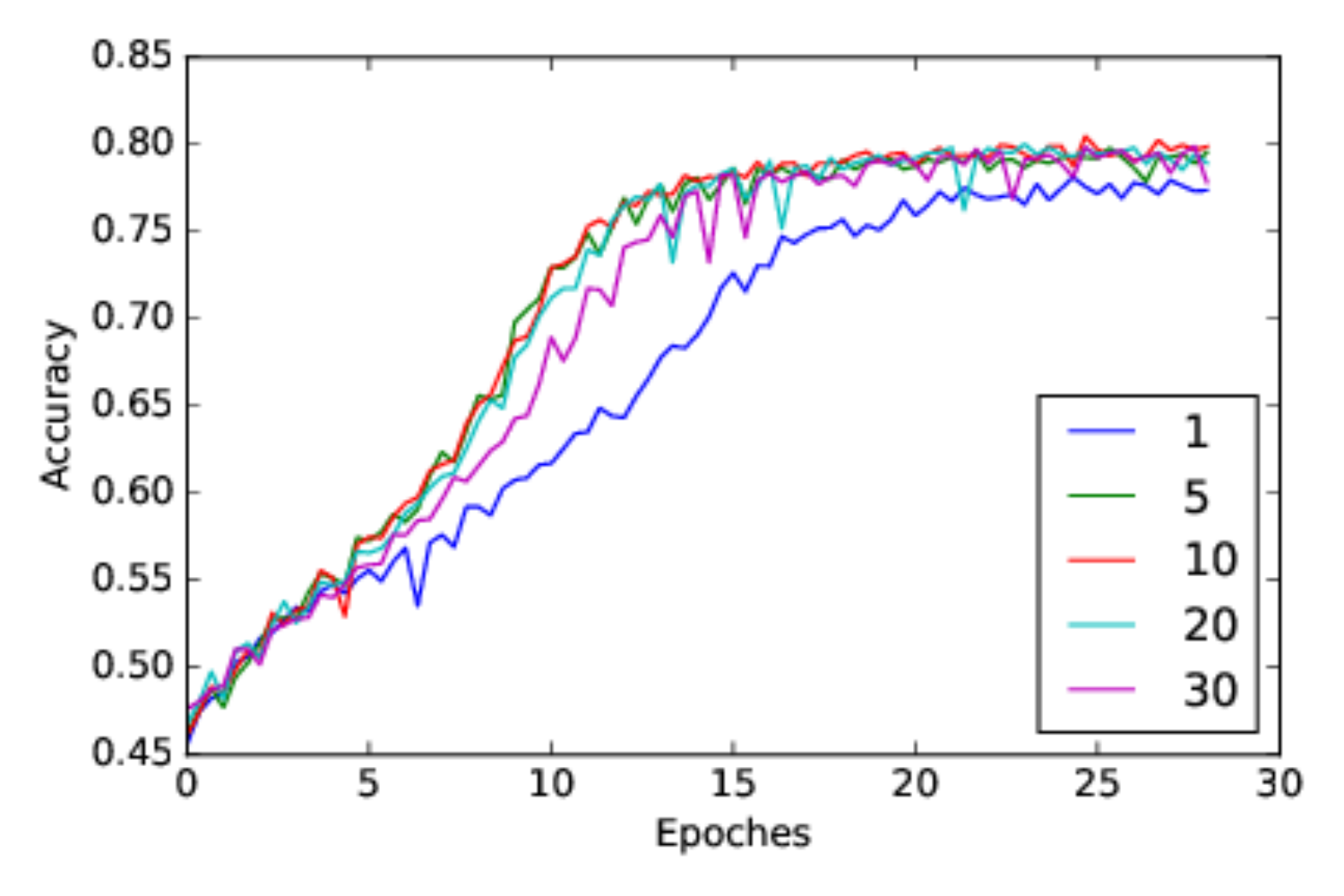}
        \caption{}
        \label{r_age}
    \end{subfigure}
    ~ 
    \begin{subfigure}[]{0.49\textwidth}
        \includegraphics[width=\linewidth]{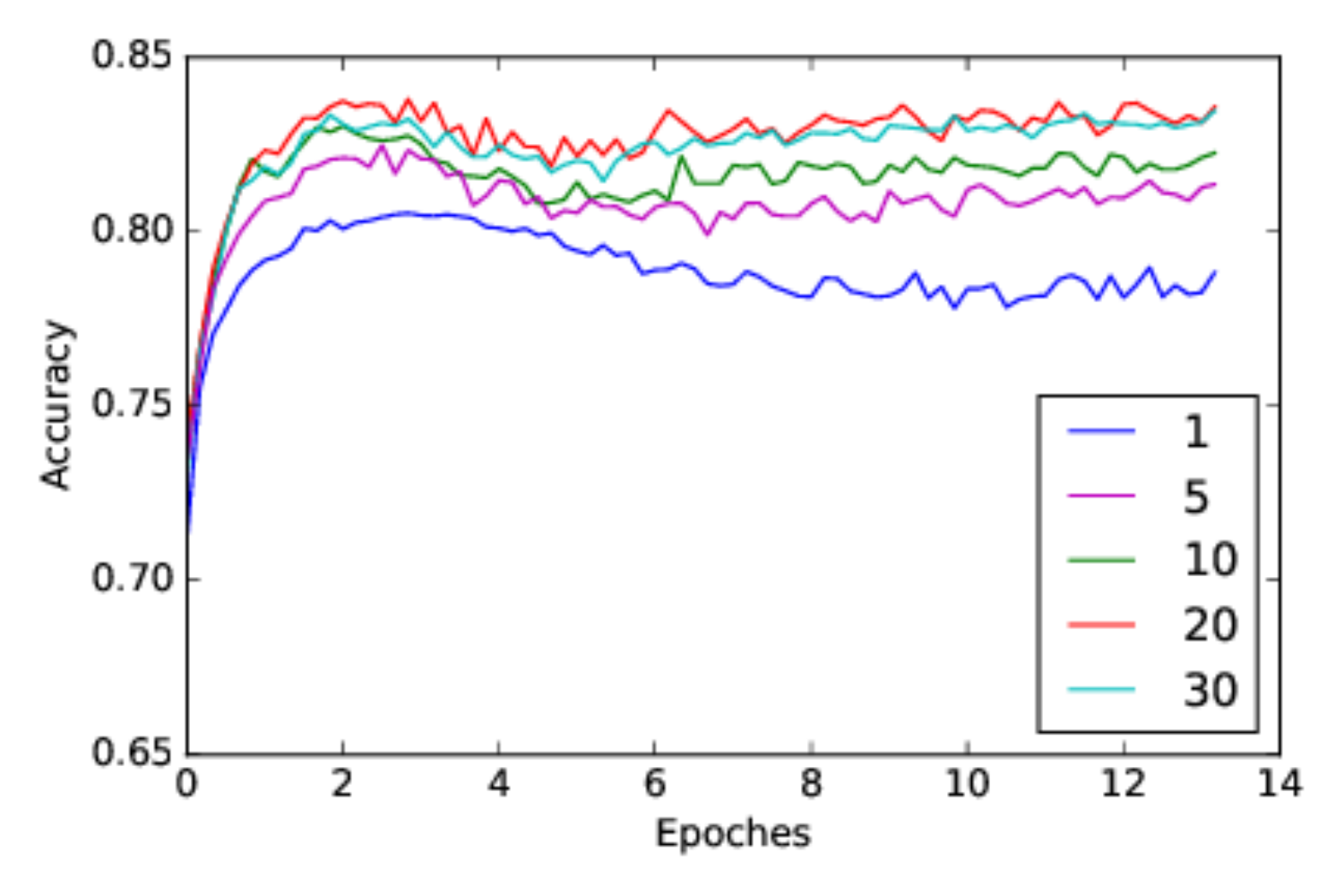}
        \caption{}
        \label{r_snli}
    \end{subfigure}
    \caption{Effect of the number of rows ($r$) in matrix sentence embedding. The vertical axes indicates test set accuracy and the horizontal axes indicates training epoches. Numbers in the legends stand for the corresponding values of $r$. (a) is conducted in Age dataset and (b) is conducted in SNLI dataset.}
    \label{varyr}
\end{figure}

Having multiple rows in the sentence embedding is expected to provide more abundant information about the encoded content. It makes sence to evaluate how significant the improvement can be brought by $r$. Taking the models we used for Age and SNLI dataset as an example, we vary $r$ from $1$ to $30$ for each task, and train the resulting $10$ models independently (Figure \ref{varyr}). Note that when $r=1$, the sentence embedding reduces to a normal vector form.

From this figure we can find that, without having multiple rows, the model performs on-par with its competitiors which use other forms of vector sentence embeddings. But there is significant difference between having only one vector for the sentence embedding and multiple vectors. The models are also quite invariant with respect to $r$, since in the two figures a wide range of values between $10$ to $30$ are all generating comparable curves.

\section{Conclusion and discussion}
In this paper, we introduced a fixed size, matrix sentence embedding with a
self-attention mechanism. Because of this attention mechanism, there is a way
to interpret the sentence embedding in depth in our model. Experimental results
over 3 different tasks show that the model outperforms other sentence embedding
models by a significant margin.

Introducing attention mechanism allows the final sentence embedding to directly
access previous LSTM hidden states via the attention summation. Thus the LSTM
doesn't need to carry every piece of information towards its last hidden state.
Instead, each LSTM hidden state is only expected to provide shorter term
context information around each word, while the higher level semantics, which
requires longer term dependency, can be picked up directly by the attention
mechanism. This setting reliefs the burden of LSTM to carry on long term
dependencies. Our experiments also support that, as we observed that our model
has a bigger advantage when the contents are longer. Further more, the notion
of summing up elements in the attention mechanism is very primitive, it can be
something more complex than that, which will allow more operations on the
hidden states of LSTM.

The model is able to encode any sequence with variable length into a fixed size
representation, without suffering from long-term dependency problems. This
brings a lot of scalability to the model: without any modification, it can be
applied directly to longer contents like paragraphs, articles, etc. Though this
is beyond the focus of this paper, it remains an interesting direction to
explore as a future work.

As a downside of our proposed model, the current training method heavily relies
on downstream applications, thus we are not able to train it in an unsupervised
way. The major obstacle towards enabling unsupervised learning in this model is
that during decoding, we don't know as prior how the different rows in the
embedding should be divided and reorganized. Exploring all those possible
divisions by using a neural network could easily end up with overfitting.
Although we can still do unsupervised learning on the proposed model by using a
sequential decoder on top of the sentence embedding, it merits more to find
some other structures as a decoder.

\subsubsection*{Acknowledgments}
The authors would like to acknowledge the developers of Theano
\citep{2016arXiv160502688short} and Lasagne. The first author would also like
to thank IBM Watson for providing resources, fundings and valuable discussions
to make this project possible, and Caglar Gulcehre for helpful discussions.

\bibliography{iclr2017_conference}
\bibliographystyle{iclr2017_conference}

\section*{APPENDIX}
\appendix

\section{Pruned MLP for Structured Matrix Sentence Embedding}  \label{appena}
As a side effect of having multiple vectors to represent a sentence, the matrix
sentence embedding is usually several times larger than vector sentence
embeddings. This results in needing more parameters in the subsequent fully
connected layer, which connects every hidden units to every units in the matrix
sentence embedding. Actually in the example shown in Figure
\ref{model_structure}, this fully connected layer takes around 90\% percent of
the parameters. See Table \ref{yelpprune}. In this appendix we are going to introduce a
weight pruning method which, by utilizing the 2D structure of matrix embedding,
is able to drastically reduce the number of parameters in the fully connected
hidden layer. 

Inheriting the notation used in the main paper, let the matrix embedding $M$
has a shape of $r$ by $u$, and let the fully connected hidden layer has $b$
units. The normal fully connected hidden layer will require each hidden unit to
be connected to every unit in the matrix embedding, as shown in Figure
\ref{model_structure}. This ends up with $r \times u \times b$ parameters in total.

However there are 2-D structures in the matrix embedding, which we should make
use of. Each row ($m_i$ in Figure \ref{model_structure}) in the matrix is
computed from a weighted sum of LSTM hidden states, which means they share some
similarities 

\begin{figure}[!b]
    \centering
    \includegraphics[width=0.8\textwidth]{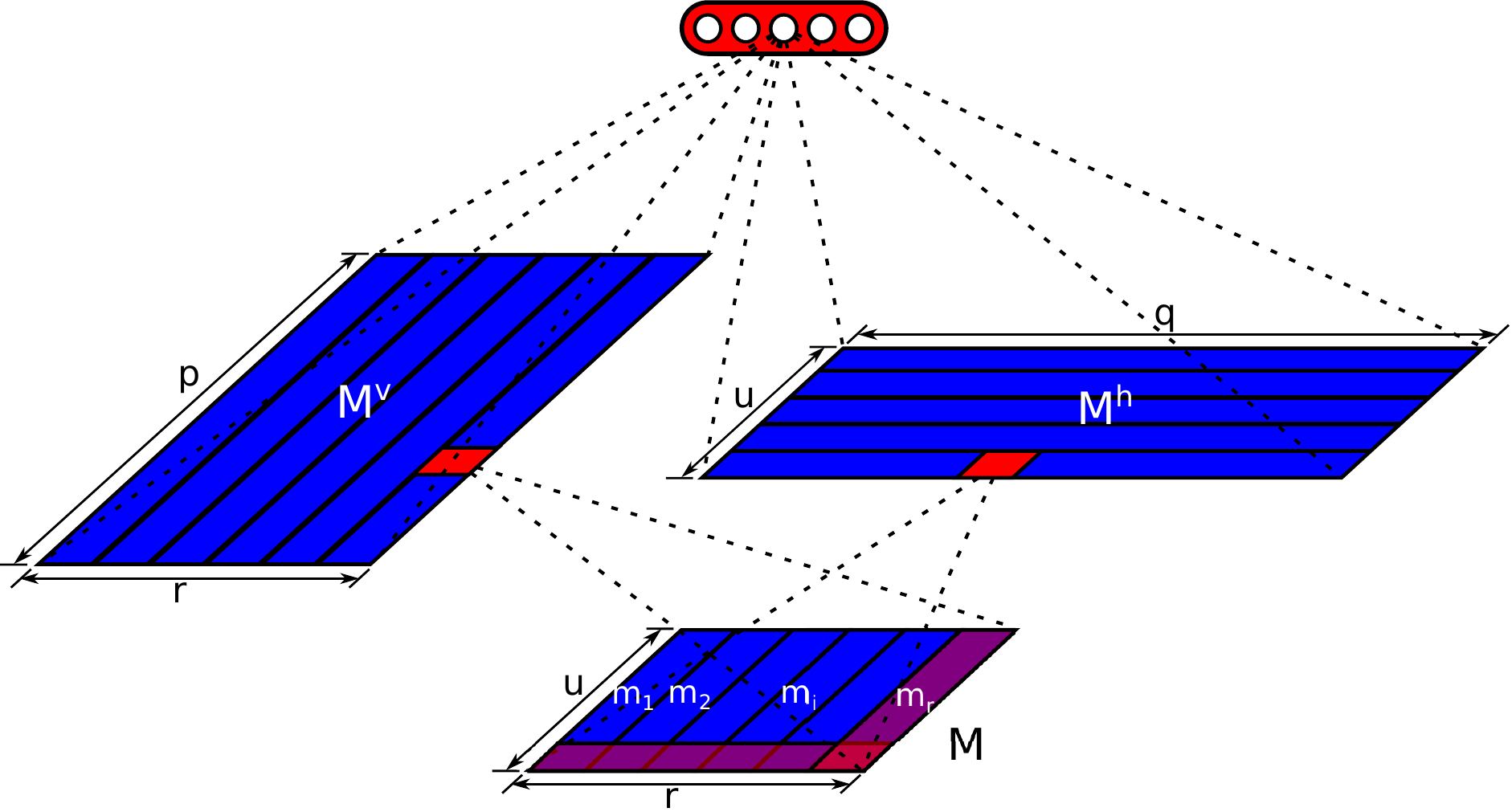}
    \caption{Hidden layer with pruned weight connections. $M$ is the matrix
    sentence embedding, $M^v$ and $M^h$ are the structured hidden representation computed by pruned weights.}
    \label{pruned_model}
\end{figure}

\begin{table}[!b]
\caption{Model Size Comparison Before and After Pruning}
\label{yelpprune}
\begin{center}
\begin{tabular}{l|lll|l|l}
    & {\bf Hidden layer}  &{\bf Softmax} &{\bf Other Parts} &{\bf Total}  &{\bf Accuracy} \\
\hline
Yelp, Original, $b$=3000       & 54M    & 15K    & 1.3M   & 55.3M  & 64.21\%          \\
Yelp, Pruned, $p$=150, $q$=10  & 2.7M   & 52.5K  & 1.3M   & 4.1M   & 63.86\%          \\
\hline
Age, Original, $b$=4000        & 72M    & 20K    & 1.3M   & 73.2M  & 80.45\%          \\
Age, Pruned, $p$=25, $q$=20    & 822K   & 63.75K & 1.3M   & 2.1M   & 77.32\%          \\
\hline
SNLI, Original, $b$=4000       & 72M    & 12K    & 22.9M  & 95.0M  & 84.43\%          \\
SNLI, Pruned, $p$=300, $q$=10  & 5.6M   & 45K    & 22.9M  & 28.6M  & 83.16\%          \\
\end{tabular}
\end{center}
\end{table}

To reflect these similarity in the fully connected layer, we split the hidden
states into $r$ equally sized groups, with each group having $p$ units. The
$i$-th group is only fully connected to the $i$-th row in the matrix
representation. All connections that connects the $i$-th group hidden units to
other rows of the matrix are pruned away. In this way, Simillarity between
different rows of matrix embedding are reflected as symmetry of connecting type
in the hidden layer. As a result, the hidden layer can be interperated as also
having a 2-D structute, with the number ($r$) and size ($p$) of groups as its
two dimensions (The $M^v$ in Figure \ref{pruned_model}). When the total number
of hidden units are the same (i.e., $r \times p = b$), this process prunes away
$(r-1)/r$ of weight values, which is a fairly large portion when $r$ is large.

On the other dimension, another form of similarity exists too. For each vector
representation $m_i$ in $M$, the $j$-th element $m_{ij}$ is a weighted sum of
an LSTM hidden unit at different time steps. And for a certain $j$-th element
in all vector representations, they are summed up from a same LSTM hidden unit.
We can also reflect this similarity into the symmetry of weight connections by
using the same pruning method we did above. Thus we will have another 2-D
structured hidden states sized $u$-by-$q$, noted as $M^h$ in Figure
\ref{pruned_model}.

Table \ref{yelpprune} takes the model we use for yelp dataset as a concrete example, and
compared the number of parameters in each part of the model, both before and
after pruning. We can see the above pruning method drastically reduces the
model size. Note that the $p$ and $q$ in this structure can be adjusted freely
as hyperparameters. Also, we can continue the corresponding pruning process on
top of $M^v$ and $M^h$ over and over again, and end up with having a stack of
structured hidden layers, just like stacking fully connected layers.

The subsequent softmax layer will be fully connected to both $M_v$ and $M_h$,
i.e., each unit in the softmax layer is connected to all units in $M_v$ and
$M_h$. This is not a problem since the speed of softmax is largely dependent of
the number of softmax units, which is not changed.In addition, for applications
like sentiment analysis and textural entailment, the softmax layer is so tiny
that only contains several units. 

Experimental results in the three datasets has shown that, this pruning
mechanism lowers performances a bit, but still allows all three models to
perform comparable or better than other models compared in the paper.

\section{Detailed Structure of the Model for SNLI Dataset}   \label{appenb}

\begin{figure}[!b]
    \centering
    \includegraphics[width=0.8\textwidth]{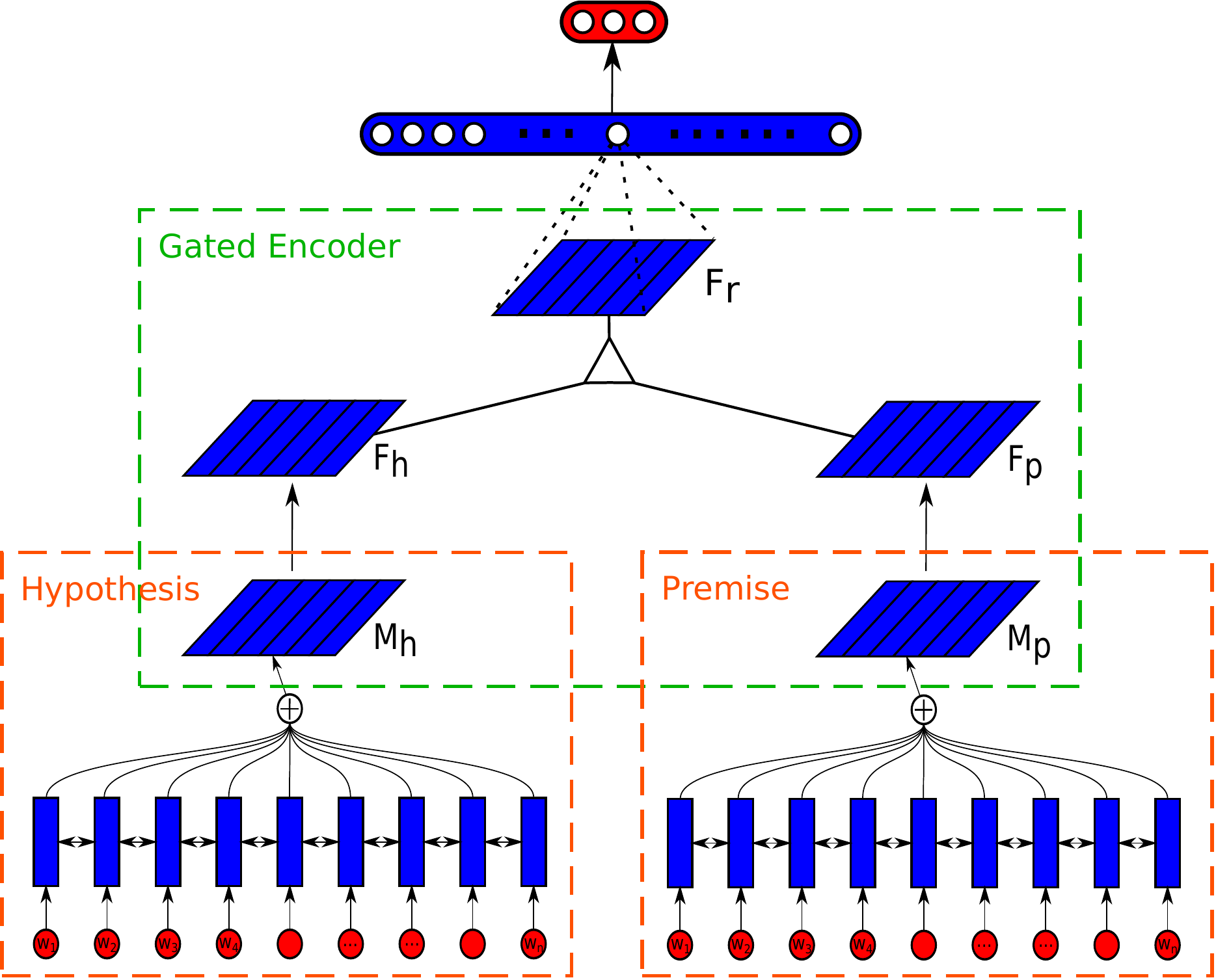}
    \caption{Model structure used for textual entailment task.}
    \label{snli_model}
\end{figure}

In Section \ref{snli} we tested our matrix sentence embedding model for the
textual entailment task on the SNLI dataset. Different from the former two
tasks, the textual entailment task consists of a pair of sentences as input. We
propose to use a set of multiplicative interactions to combine the two matrix
embeddings extracted for each sentence. The form of multiplicative interaction
is inspired by \emph{Factored Gated Autoencoder} \citep{memisevic2013learning}.

The overall structure of our model for SNLI is dipicted in Figure
\ref{snli_model}. For both hypothesis and premise, we extract their embeddings
($M_h$ and $M_p$ in the figure) independently, with a same LSTM and attention
mechanism. The parameters of this part of model are shared (rectangles with
dashed orange line in the figure).

Comparing the two matrix embeddings corresponds to the green dashed rectangle
part in the figure, which computes a single matrix embedding ($F_r$) as the
factor of semantic relation between the two sentences. To represent the relation between
$M_h$ and $M_p$, $F_r$ can be connected to $M_h$ and $M_p$ through a 
three-way \emph{multiplicative interaction}. In a three-way multiplicative
interaction, the value of anyone of $F_r$, $M_h$ and $M_p$ is a function of the
product of the others. This type of connection is originally introduced to extract
relation between images \citep{memisevic2013learning}. Since here we are
just computing the factor of relations ($F_r$) from $M_h$ and $M_p$, it
corresponds to the encoder part in
the Factored Gated Autoencoder in \cite{memisevic2013learning}. We call it
Gated Encoder in Figure \ref{snli_model}.

First we multiply each row in the matrix embedding by a different weight
matrix. Repeating it over all rows, corresponds to a batched dot product
between a 2-D matrix and a 3-D weight tensor. Inheriting the name in
\citep{memisevic2013learning}, we call the resulting matrix as \emph{factor}.
Doing the batched dot for both hypothesis embedding and premise embedding, we
have $F_h$ and $F_p$, respectively.

\begin{equation}
    F_h = batcheddot(M_h, W_{fh})
\end{equation}
\begin{equation}
    F_p = batcheddot(M_p, W_{fp})
\end{equation}

Here $W_{fh}$ and $W_{fp}$ are the two weight tensors for hypothesis embedding
and premise embedding.

The factor of the relation ($F_r$) is just an element-wise
product of $F_h$ and $F_p$ (the triangle in the middle of Figure \ref{snli_model}):

\begin{equation}
    F_r = F_h \odot F_p
\end{equation}

Here $\odot$ stands for element-wise product. After the $F_r$ layer, we then use
an MLP with softmax output to classify the relation into different categlories.

\appendix

\end{document}